\pdfoutput=1

\PassOptionsToPackage{table}{xcolor}
\documentclass[11pt]{article}

\usepackage[]{acl}

\usepackage{times}
\usepackage{latexsym}
\usepackage{xspace}
\usepackage{fontawesome}
\usepackage{inconsolata}
\usepackage[T1]{fontenc}

\usepackage{tcolorbox}
\usepackage{xcolor}

\definecolor{mygreen}{RGB}{10,150,10}
\definecolor{myblue}{RGB}{10,10,150}
\definecolor{myred}{RGB}{150,10,10}
\definecolor{mypink}{RGB}{255,242,175}
\definecolor{mygray}{RGB}{0,0,0}

\definecolor{darkgreen}{RGB}{3, 75, 3}
\newcommand{\modelsub}[1]{\textsubscript{{\fontfamily{lmr}\selectfont \textit{#1}}}}

\newcommand{\stitle}[1]{\vspace{1ex} \noindent{\bf #1}}

\newcommand{\modelname}{\modelnamens\xspace}
\newcommand{\modelnamens}{\textsc{SpaBERT}}
\newcommand{\redloc}{{\color{myred}{\faMapMarker}\xspace}}
\newcommand{\greenloc}{{\color{mygreen}{\faMapMarker}} \xspace}
\newcommand{\blueloc}{{\color{myblue}{\faMapMarker}}\xspace}

\newtcbox{\highlight}[0]{boxsep=0pt,left=2pt,top=2pt,bottom=1pt,right=2pt,boxrule=0.4pt,arc=2pt,auto outer arc,colback=mypink,colframe=mygray,enlarge bottom by=-3.5pt,fonttitle=\texttt}
\newcommand{\speedway}{\mbox{\highlight{\texttt{Speedway}}}\xspace}
\newcommand{\speedwaywithred}{[\mbox{\highlight{\texttt{Speedway}}}\texttt{,}{\color{myred}{\faMapMarker}}]}
\newcommand{\speedwaywithgreen}{[\mbox{\highlight{\texttt{Speedway}}}\texttt{,}{\color{mygreen}{\faMapMarker}}]}
\newcommand{\speedwaywithblue}{[\mbox{\highlight{\texttt{Speedway}}}\texttt{,}{\color{myblue}{\faMapMarker}}]}

\newcommand{\logo}{\includegraphics[height=1em]{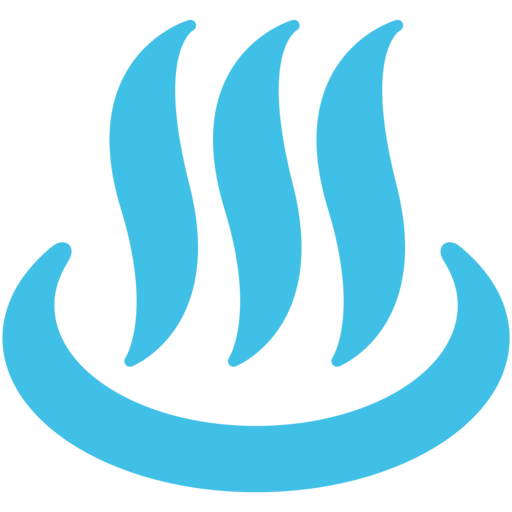}}

\newcommand{\tablesupport}{
\begin{table}
\centering
\small
\begin{tabular}{lrr}
\hline 
\textbf{Classes} & \textbf{California} & \textbf{London} \\ \hline

Education                   & 6,222 & 618     \\
Entertainment\_Arts\_Culture  &1,380 &  601    \\
Facilities                  & 574 &  179    \\
Financial                   & 2,590&  769     \\
Healthcare                  & 3,779&  1,779   \\
Public\_Service             & 2,658 &   393   \\
Sustenance                 & 4,276 &  1,693 \\
Transportation             & 4,226 &  1,618   \\
Waste\_Management         & 167 &   76   \\
\hline
\textbf{Total}  & 25,872 & 7,726 \\
\hline 
\end{tabular}
\vspace{-0.5em}
\caption{\label{tab:typing_data_support} OSM Geo-entity typing dataset statistics. The first column shows the nine OSM semantic types, following by the numbers of samples for each class in California and London.}
\vspace{-0.5em}
\end{table}
}

\newcommand{\tabletypingresults}{
\begin{table*}
\centering
\small
\begin{tabular}{lcccccccccc}
\hline 
\textbf{Classes $\rightarrow$} & \textbf{Edu.} & \textbf{Ent.}& \textbf{Fac.}& \textbf{Fin.}& \textbf{Hea.}& \textbf{Pub.}& \textbf{Sus.}& \textbf{Tra.}& \textbf{Was.} & \textbf{Micro Avg} \\ \hline

BERT\modelsub{Base} &  \underline{.674} &   \underline{.634} &   \underline{\textbf{.763}} &   .929 &   .856 &   .872 &   .856 &   .862 &   .678 & \underline{.835} \\
RoBERTa\modelsub{Base} & .626 &   .627 &   .605 &   \underline{.951} &   \underline{\textbf{.869}} &   .818 &   .838 &   .850 &   .475 & .820 \\
SpanBERT\modelsub{Base}	&   .633 &   .589 &   .608 &   .916 &   .859 &   \underline{.882} &   .824 &   \underline{.867} &   \underline{\textbf{.735}} & .819 \\
LUKE\modelsub{Base} &  .648 &   .608 &   .598 &   .945 &   .857 &   .867 &   .854 &   .851 &   .517 & .825 \\
SimCSE\modelsub{BERT-Base} & .623 &   .590 &   .504 &   .925 &   .867 &   .852 &   \underline{.857} &   .810 &   .470 & .810 \\
SimCSE\modelsub{RoBERTa-Base} &   .621 &   .629 &   .499 &   \underline{.951} &   .841 &   .853 &   .828 &   .856 &   .500 & .814 \\
\hline
\cellcolor{blue!5}\modelnamens\modelsub{Base} &  \cellcolor{blue!5}\textbf{.674} &   \cellcolor{blue!5}\textbf{.653} &  \cellcolor{blue!5}.680 &   \cellcolor{blue!5}\textbf{.959} &   \cellcolor{blue!5}.865 &   \cellcolor{blue!5}\textbf{.900} &   \cellcolor{blue!5}\textbf{.883} &   \cellcolor{blue!5}\textbf{.888} &   \cellcolor{blue!5}.703 & \cellcolor{blue!5}\textbf{.852} \\
\hline \hline
BERT\modelsub{Large} &  \underline{.707} &   \underline{.661} &   .647 &   .937 &   .874 &   .850 &   \underline{.873} &   .864 &   .526 & .841 \\
RoBERTa\modelsub{Large} &  .657 &   .626 &   .682 &   .907 &   .855 &   .805 &   .831 &   .859 &   .587 & .817 \\
SpanBERT\modelsub{Large} &  .683 &   .652 &   .661 &   .931 &   .868 &   .853 &   .851 &   .848 &   \underline{.624} & .829 \\
LUKE\modelsub{Large} &  .665 &   .607 &   .660 &   .899 &   .855 &   .809 &   .813 &   .844 &   .587 & .808 \\
SimCSE\modelsub{BERT-Large} &  .693 &   \underline{.661} &   \underline{\textbf{.713}} &   \underline{.940} &   \underline{.880} &   \underline{.871} &   .864 &   \underline{.867} &   .564 & \underline{.844} \\
SimCSE\modelsub{RoBERTa-Large} &  .683 &   .630 &   .648 &   .916 &   .865 &   .802 &   .807 &   .848 &   .587 & .811 \\
\hline

\cellcolor{blue!5}\modelnamens\modelsub{Large} &   \cellcolor{blue!5}\textbf{.731} &   \cellcolor{blue!5}\textbf{.690} &   \cellcolor{blue!5}.710 &   \cellcolor{blue!5}\textbf{.956} &   \cellcolor{blue!5}\textbf{.901} &   \cellcolor{blue!5}\textbf{.892} &   \cellcolor{blue!5}\textbf{.893} &   \cellcolor{blue!5}\textbf{.903} &   \cellcolor{blue!5}\textbf{.677} & \cellcolor{blue!5}\textbf{.871} \\

\hline

\end{tabular}
\vspace{-0.5em}
\caption{\label{tab:typing_result} Geo-entity typing with the state-of-the-art LMs and \modelname. Column names are the OSM classes. Bold numbers are the highest scores in each column. Underlined  numbers are the highest scores among baselines.}
\vspace{-0.5em}
\end{table*}
}

\newcommand{\tablelinkingresults}{
\begin{table}
\centering
\resizebox{1.0\columnwidth}{!}{
\begin{tabular}{lccccc}
\hline \textbf{Model} & \textbf{MRR} & \textbf{R@1} &  \textbf{R@5} & \textbf{R@10} \\ \hline

BERT\modelsub{Base}	& .400	& .289		& \underline{.559}	& \underline{.635} \\
RoBERTa\modelsub{Base}	& .326	& .232		& .446	& .540 \\
SpanBERT\modelsub{Base}	& .164	& .138		& .201	& .213 \\
LUKE\modelsub{Base}	& .306	& .188		& .440	& .547 \\
SimCSE\modelsub{BERT-Base}	& \underline{.453}	& \underline{\textbf{.371}}		& .547	& .628 \\
SimCSE\modelsub{RoBERTa-Base}	& .227	& .188		& .264	& .301 \\
\hline 
\cellcolor{blue!5}\modelnamens\modelsub{Base} & \cellcolor{blue!5}\textbf{.515} & \cellcolor{blue!5}.338& \cellcolor{blue!5}\textbf{.744}& \cellcolor{blue!5}\textbf{.850}\\
\hline \hline
BERT\modelsub{Large}	& .337	& .245		& .459	& .509 \\
RoBERTa\modelsub{Large}	& .379	& .220		& .603	& .704 \\
SpanBERT\modelsub{Large}	& .229	& .176		& .308	& .339 \\

LUKE\modelsub{Large}	& .402	& .232		& \underline{.635}	& \underline{.767} \\
SimCSE\modelsub{BERT-Large}	& \underline{.475}	& \underline{\textbf{.402}}		& .559	& .616 \\
SimCSE\modelsub{RoBERTa-Large}	& .214	& .176		& .239	& .283 \\
\hline 

\cellcolor{blue!5}\modelnamens\modelsub{Large} & \cellcolor{blue!5}\textbf{.537} & \cellcolor{blue!5}.383& \cellcolor{blue!5}\textbf{.744}&  \cellcolor{blue!5}\textbf{.864}\\

\hline
\end{tabular}
}
\vspace{-0.5em}
\caption{\label{tab:linking_result} Geo-entity linking result. Bold and underlined numbers are the highest scores in each column and the highest scores among the baselines, respectively. }
\vspace{-0.5em}
\end{table}
}

\newcommand{\tablelinkingscaleresults}{
\begin{table}[t]
\centering
\resizebox{\columnwidth}{!}{
\begin{tabular}{crcccc}
\hline \textbf{Map Set} & \textbf{Scale $\Downarrow$} & \textbf{MRR} & \textbf{R@1} &  \textbf{R@5} & \textbf{R@10} \\ \hline
15-CA & 1:62500  &0.503  & 0.422 & 0.680 & 0.914  \\
30-CA &	1:125000 &0.639  & 0.599	& 0.862	& 0.932 \\
60-CA &	1:250000 &0.404  & 0.133	& 0.678	& 0.800 \\
\hline
\end{tabular}
}
\vspace{-0.5em}
\caption{\label{tab:linking_result_scale} Impact of entity omission for linking (USGS and Wikidata). Results from \modelnamens\modelsub{Large}. 1:625000 means 1 centimeter on a map represents 625 meters in the physical world. }
\vspace{-1em}
\end{table}
}

\newcommand{\tabletypingablation}{
\begin{table}[t]
\centering
\resizebox{\columnwidth}{!}{
\setlength\tabcolsep{2pt}
\begin{tabular}{cccccccccc}
\hline 
\textbf{\#Neighbor}& \textbf{1} & \textbf{2} & \textbf{3} & \textbf{4} & \textbf{5} & \textbf{6}  & \textbf{7} & \textbf{8} & \textbf{9} \\ \hline

\textit{Base} & .773 & .808 & .814 & .827 & .835 & .831 & .835 & .836 & .840\\

\textit{Large} & .795 & .822 & .834 & .838 & .843 & .847 & .852 & .848 & .854\\
\hline

\textbf{\#Neighbor}& \textbf{10} &\textbf{20 }& \textbf{30} & \textbf{40} & \textbf{50} & \textbf{60} & \textbf{70} & \textbf{80} & \textbf{90} \\
\hline
\textit{Base} & .843 & .844 & .846 & .851 & .852 & .850 & .849 & .851 & .852\\
\textit{Large} &.850 &.857 &.862 &.858 &.858 &.868 &.863 &.867 &.869 \\

\hline
\end{tabular}
}
\vspace{-0.5em}
\caption{\label{tab:typing_ablation} Impact of the pseudo sentence length. Numbers in the table are micro-F1 scores on the OSM typing dataset. \textit{Base} and \textit{Large} are short for \modelnamens\modelsub{Base} and \modelnamens\modelsub{Large}. }
\vspace{-0.5em}
\end{table}
}

\usepackage[T1]{fontenc}

\usepackage[utf8]{inputenc}

\usepackage{microtype}

\usepackage{comment}
\usepackage{multirow}
\usepackage{booktabs}
\usepackage{amsfonts}
\usepackage{color,xcolor}
\usepackage{amsmath}
\usepackage{graphicx}
\usepackage{setspace}
\usepackage{diagbox}
\usepackage{amssymb}
\usepackage{rotating}
\usepackage{array}
\usepackage{epstopdf}
\usepackage{tikz}

\usepackage{cleveref}
\crefformat{section}{\S#2#1#3}
\crefformat{subsection}{\S#2#1#3}
\crefformat{subsubsection}{\S#2#1#3}
\crefrangeformat{section}{\S\S#3#1#4 to~#5#2#6}
\crefmultiformat{section}{\S\S#2#1#3}{ and~#2#1#3}{, #2#1#3}{ and~#2#1#3}
\usepackage{refstyle}
\Crefformat{figure}{#2Fig.~#1#3}
\Crefmultiformat{figure}{Figs.~#2#1#3}{ and~#2#1#3}{, #2#1#3}{ and~#2#1#3}
\Crefformat{table}{#2Tab.~#1#3}
\Crefmultiformat{table}{Tabs.~#2#1#3}{ and~#2#1#3}{, #2#1#3}{ and~#2#1#3}
\Crefformat{appendix}{Appx.~\S#2#1#3}
\crefformat{algorithm}{Alg.~#2#1#3}

\title{\modelname: 
A Pretrained Language Model from Geographic Data for Geo-Entity Representation
}

\author{Zekun Li$^1$, Jina Kim$^1$, Yao-Yi Chiang$^1$, Muhao Chen$^2$ \\
        $^1$Department of Computer Science and Engineering, University of Minnesota, Twin Cities\\
        $^2$Department of Computer Science, University of Southern California \\
        \texttt{\{li002666,kim01479,yaoyi\}@umn.edu, muhaoche@usc.edu}
}

\begin{document}
\maketitle
\begin{abstract}

Named geographic entities (geo-entities for short)
are the building blocks of many geographic datasets. Characterizing geo-entities is integral to various application domains, such as geo-intelligence and map comprehension, while a key challenge is to capture the spatial-varying context of an entity. We hypothesize that we shall know the characteristics of a geo-entity by its surrounding entities, similar to knowing word meanings by their linguistic context. Accordingly, we propose a novel spatial language model, \modelname (\logo), which provides a general-purpose geo-entity representation based on neighboring entities in geospatial data. \modelname extends BERT to capture linearized spatial context, while incorporating a spatial coordinate embedding mechanism to preserve spatial relations of entities in the 2-dimensional space. \modelname is pretrained with masked language modeling and masked entity prediction tasks to learn spatial dependencies. We apply \modelname to two downstream tasks: geo-entity typing and geo-entity linking.  Compared with the existing language models that do not use spatial context, \modelname shows significant performance improvement on both tasks. We also analyze the entity representation from \modelname in various settings and the effect of spatial coordinate embedding.


\end{abstract}

\section{Introduction}

\begin{figure*}[t]
  \begin{center}
	\includegraphics[width=1.0\linewidth]{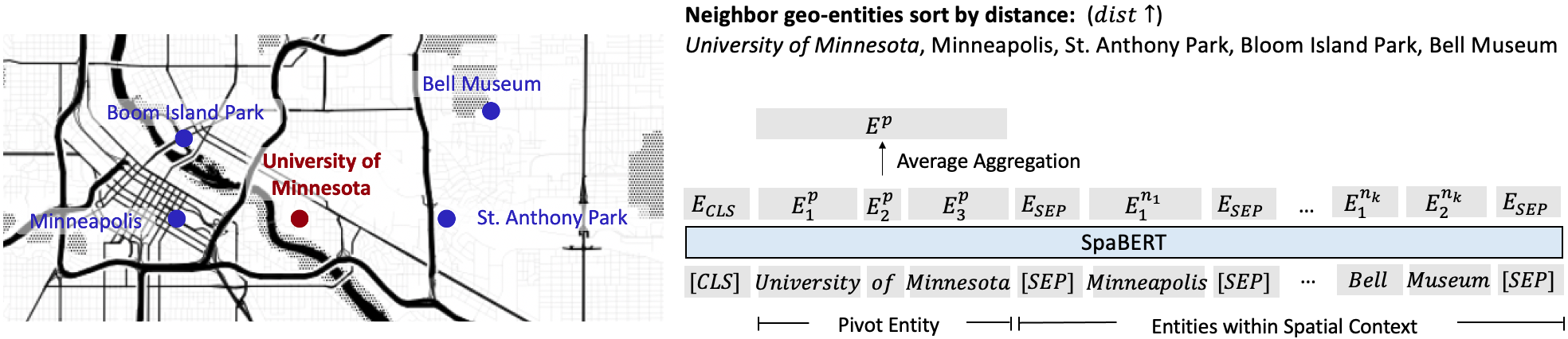}
  \end{center}
  \caption{Overview for generating the pivot entity representation, $E^p$ (\tikz\draw[myred,fill=myred] (0,0) circle (.5ex);: pivot entity; \tikz\draw[myblue,fill=myblue] (0,0) circle (.5ex);: neighboring geo-entities). \modelname sorts and concatenate neighboring geo-entities by their distance to pivot in ascending order to form a pseudo-sentence. \texttt{[CLS]} is prepended at the beginning. \texttt{[SEP]} separates entities. \modelname generates token representations and aggregates representations of pivot tokens to produce $E^p$.}\label{fig:sbert_outline}
\end{figure*}

Interpreting human behaviors requires considering human activities and their surrounding environment. Looking at a stopping location, \speedwaywithred,\footnote{A geographic entity name \speedway and its location~\redloc~(e.g., latitude and longitude). Best viewed in color.} ~from a person's trajectory, we might assume that this person needs to use the location's amenities if \speedway implies a gas station, and \redloc~is near a highway exit. We might predict a meetup at \speedwaywithred~ if the trajectory travels through many other locations, \blueloc, \greenloc, ..., of the same name, \speedway, to arrive at \speedwaywithred~ in the middle of farmlands. As humans, we are able to make such inferences using the name of a geographic entity (geo-entity) and other entities in a spatial neighborhood. Specifically, we contextualize a geo-entity by a reasonable surrounding neighborhood learned from experience and, from the neighborhood, relate other relevant geo-entities based on their name and spatial relations (e.g., distance) to the geo-entity. This way, even if two gas stations have the same name (e.g., \speedway ) and entity type (e.g., `gas station'),  we can still reason about their spatially varying semantics and use the semantics for prediction.

Capturing this spatially varying location semantics can help 
recognizing and resolving geospatial concepts (e.g., toponym detection, typing and linking) and the grounding of geo-entities in documents, 
scanned historical maps, and a variety of knowledge bases, such as Wikidata, OpenStreetMap, and GeoNames. Also, the location semantics can support effective use of spatial textual information (geo-entities names) in many spatial computing task, including moving behavior detection from visiting locations of trajectories~\cite{Yue2021-aa,Yue2019-hn}, point of interest recommendations~\cite{yin2017spatial,zhao2020go}, air quality~\cite{Lin2017-go,Lin2018-xw, Lin2020-kk,jiang-etal-2019-enhancing} and traffic prediction~\cite{yuan2021survey, Gao2019-fo} using location context.  

Recently, the research community has seen a rapid advancement in pretrained language models (PLMs) \cite{devlin-etal-2019-bert,liu2019roberta, lewis-etal-2020-bart, sanh2019distilbert}, which supports strong contextualized language representation abilities \cite{DBLP:conf/iclr/LanCGGSS20} and serves as the backbones of various NLP systems  \cite{rothe2020leveraging, yang-etal-2019-exploring-pre}. The extensions of these PLMs help NL tasks in different data domains (e.g., biomedicine \cite{lee2020biobert,phan2021scifive}, software engineering \cite{tabassum-etal-2020-code}, finance \cite{liu2021finbert}) and modalities (e.g., tables \cite{herzig-etal-2020-tapas,yin-etal-2020-tabert,wang-etal-2022-robust} and images \cite{li-etal-2020-bert-vision,su2019vl}). However, it is challenging to adopt existing PLMs or their extensions to capture geo-entities' spatially varying semantics. First, geo-entities exist in the physical world. Their spatial relations  (i.e., distance and orientation) do not have a fixed structure (e.g., within a table or along a row of a table) that can help contextualization. Second, existing language models (LMs) pretrained on general domain corpora~\cite{devlin-etal-2019-bert,liu2019roberta} require fine-tuning for domain adaptation to handle names of geo-entities.

To tackle these challenges, we present \modelname (\logo), a LM that captures the spatially varying semantics of geo-entity names using large geographic datasets for entity representation. Built upon BERT \cite{devlin-etal-2019-bert}, \modelname generates the contextualized representation for a geo-entity of interest (referred to as the \textit{pivot}), based on its geographically nearby geo-entities. Specifically, \modelname linearizes the 2-dimensional spatial context by forming \textit{pseudo sentences} that consist of names of the pivot and neighboring geo-entities, ordered by their spatial distance to the \textit{pivot}. \modelname also encodes the spatial relations between the pivot and neighboring geo-entities with a continuous \emph{spatial coordinate embedding}. The spatial coordinate embedding models horizontal and vertical distance relations separately and, in turn, can capture the orientation relations. These techniques make the inputs compatible with the BERT-family structures. 

In addition, the backbone LM, BERT, in \modelname is pretrained with general-purpose corpora and would not work well directly on geo-entity names because of the domain shift. We thus train \modelname using  \textit{pseudo sentences} generated from large geographic datasets derived from OpenStreetMap (OSM).\footnote{\label{ft:osm}OpenStreetMap: \url{https://www.openstreetmap.org/}} This pretraining process conducts Masked Language Modeling (MLM) and Masked Entity Prediction (MEP) that randomly masks subtokens and full geo-entity names in the pseudo sentences, respectively. MLM and MEP enable \modelname to learn from pseudo sentences for generating spatially varying contextualized representations for geo-entities.

\modelname provides a general-purpose representation for geo-entities based on their spatial context. Similar to linguistic context, spatial context refers to the surrounding environment of a geo-entity. For example, \speedwaywithred, \speedwaywithblue, and \speedwaywithgreen ~would have different representations since the surrounding environment of \redloc, \blueloc, and \greenloc could vary.    
We evaluate \modelname on two tasks: 1) geo-entity typing and 2) geo-entity linking to external knowledge bases. Our analysis includes the performance comparison of \modelname in various settings due to characteristics of geographic data sources (e.g., entity omission).

To summarize, this work has the following contributions. We propose an approach to linearize the 2-dimensional spatial context, encode the geo-entity spatial relations, and use a LM to produce spatial varying feature representations of geo-entities. We show that \modelname is a general-purpose encoder by supporting geo-entity typing and geo-entity linking, which are keys to effectively integrating large varieties of geographic data sources, the grounding of geo-entities as well as supports for a broad range of spatial computing applications. The experiments demonstrate that \modelname is effective for both tasks and outperforms the SOTA LMs. 

\section{\modelname}
\modelname is a LM built upon a pretrained BERT and further trained to produce contextualized geo-entity representations given large geographic datasets. \Cref{fig:sbert_outline} shows the outline of \modelname, with the details below. This section first presents the preliminary (\Cref{sec:prelim}) and then describes the overall approach  for learning geo-entities' contextualized representations (\Cref{sec:overall}), the pretraining strategies (\Cref{sec:pretrain}), and inference procedures (\Cref{sec:inference}). 
\begin{figure*}[t]
  \begin{center}
	\includegraphics[width=1.0\linewidth]{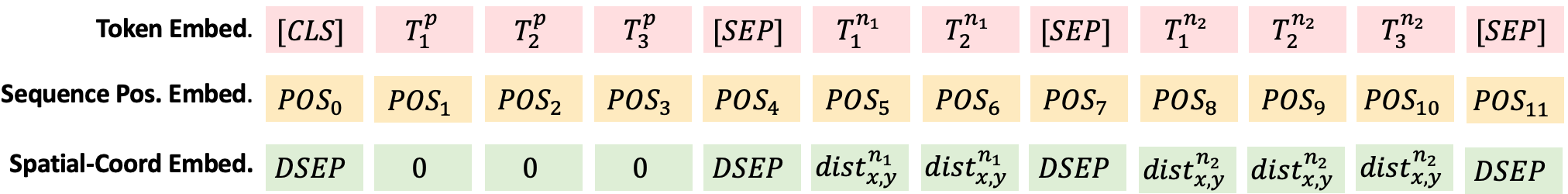}
  \end{center}
  \vspace{-0.5em}
  \caption{Embedding modules. Inputs to the token-embedding are the tokenized geo-entity names in the pseudo-sentence. Inputs to the sequence position embedding are the sequence indices of the tokens. Inputs to the spatial coordinate embedding are the normalized distances from the pivot in each spatial dimension  (details in \Cref{sec:overall}). }
  \label{fig:sbert_embedding}
  \vspace{-0.5em}
\end{figure*}
\subsection{Preliminary} \label{sec:prelim}
We assume that given a geographic dataset (e.g., OSM) with many geo-entities, $S = \{g_1, g_2, ..., g_l\}$, each geo-entity $g_i$ (e.g., \speedwaywithred) has two attributes: name $g^{name}$(\speedway) and location $g^{loc}$(\redloc). The location attribute, $g^{loc}$, is a tuple $g^{loc} = (g^{locx}, g^{locy}, ...)$ that identifies the location in a coordinate system (e.g., x and y image pixel coordinates or latitude and longitude geo-coordinates with altitudes). WLOG, here we assume a 2-dimensional space. \modelname aims to generate a contextualized representation for each geo-entity $g_i$ in $S$ given its spatial context. We denote a geo-entity that we seek to contextualize as the \emph{pivot entity}, $g_p$, $p$ for short. The spatial context of $p$ is $SC(p) = \{g_{n_1},...,g_{n_k}\}$ where $distance(p, g_{n_k}) < T$. $T$ is a spatial distance parameter defining a local area. We call $g_{n_1},...,g_{n_k}$ as $p$'s neighboring geo-entities and denote them as $n_1, ..., n_k$ when there is no confusion.
\subsection{Contextualizing Geo-entities} \label{sec:overall}

\stitle{Linearizing Neighboring Geo-entity Names} For a pivot, $p$, \modelname first linearizes its neighboring geo-entitie names to form a BERT-compatible input sequence, called a \emph{pseudo sentence}.  The corresponding pseudo sentence for the example in \Cref{fig:sbert_outline} is constructed as:

\begin{center}
\vspace{0.1em}
\noindent\fbox{%
    \parbox{0.96\linewidth}{%
     {\fontfamily{cmtt}\selectfont
     \small
        [CLS] University of Minnesota [SEP] Minneapolis [SEP] St. Anthony Park [SEP] Bloom Island Park [SEP] Bell Museum [SEP] }
    }%
}
\vspace{0.2em}
\end{center}

The pseudo sentence starts with the pivot name followed by the names of the pivot's neighboring geo-entities, ordered by their spatial distance to the pivot in ascending order. The idea is that nearby geo-entities are more related (for contextualization) than distant geo-entities. \modelname also tokenizes the pseudo sentences using the original BERT tokenizer with the special token \texttt{[SEP]} to separate entity names. The subtokens of a neighbor $n_k$ is denoted as $T^{n_k}_j$ as in \Cref{fig:sbert_embedding}.

\stitle{Encoding Spatial Relations} \modelname adopts two types of position embeddings in addtion to the token embeddings in the pseudo sentences (\Cref{fig:sbert_embedding}). The sequence position embedding represents the token order, same as the original 
position embedding in BERT and other Transformer LMs. Also, \modelname incorporates a \emph{spatial coordinate embedding} mechanism, which seeks to represent the spatial relations between the pivot and its neighboring geo-entities. \modelname's spatial coordinate embeddings encode each location dimension separately to capture the relative distance and orientation between geo-entities. Specifically, for the 2-dimensional space, \modelname generates normalized distance $dist_x^{n_k}$ and $dist_y^{n_k}$ for each neighboring geo-entity using the following equations:
\begin{align*}\label{eq:dist}
    dist_x^{n_k} = (g_{n_k}^{locx} - g_p^{locx})/Z \\
    dist_y^{n_k} = (g_{n_k}^{locy} - g_p^{locy})/Z
\end{align*}
where $Z$ is a normalization factor, and ($g_{p}^{locx}$, $g_{p}^{locy}$), ($g_{n_k}^{locx}$, $g_{n_k}^{locy}$) are the locations of pivot $p$ and neighboring entity $n_k$. Note that the tokens belong to the same geo-entity name have the same $dist_x^{n_k}$ and $dist_y^{n_k}$. Also, \modelname uses \texttt{DSEP}, a constant numerical value larger than $max(dist_x^{n_k},dist_y^{n_k})$ for all neighboring entities to differentiate special tokens from entity name tokens. 

\modelname encodes $dist_x^{n_k}$ and $dist_y^{n_k}$ using a continuous spatial coordinate embedding layer with real-valued distances as input to preserve the continuity of the output embeddings. Let $\mathcal{S}_{n_k}$ be the spatial coordinate embedding of the neighbor entity $n_k$, $M$ be the embedding's dimension, and $\mathcal{S}_{n_k} \in \mathbf{R}^M$. We define $\mathcal{S}_{n_k}$ as:
\begin{align*}
  \mathcal{S}_{n_k}^{(m)} = 
  \begin{cases}
    sin(dist^{n_k}/10000^{2j/M}), m = 2j  \\
    cos(dist^{n_k}/10000^{2j/M}), m = 2j+1 
  \end{cases}
\end{align*}
where $\mathcal{S}_{n_k}^{(m)}$ is the $m$-th component of $\mathcal{S}_{n_k}$. $dist^{n_k}$ represents $dist_x^{n_k}$ and $dist_y^{n_k}$ for the spatial coordinate embeddings along the horizontal and vertical directions, respectively.

The token embedding, sequence position embedding and spatial coordinate embedding are summed up then fed into the encoder. Similar to BERT, \modelname encoder calculates an output embedding for each token. Then \modelname averages the pivot's token-level embeddings to produce a fixed-length embedding for the pivot's contextualized representation.

\subsection{Pretraining}\label{sec:pretrain}
We train \modelname with two tasks to adapt the pretrained BERT backbone to geo-entity pseudo sentences. One task is the masked language modeling (MLM)~\cite{devlin-etal-2019-bert}, for which \modelname needs to learn how to complete the full names of geo-entities from pseudo sentences with randomly masked subtokens using the remaining subtokens and their spatial coordinates (i.e., partial names and spatial relations between subtokens). The block below shows an example of the masked input for MLM, where \#\#\# are the masked subtokens.

\begin{center}
\vspace{0.1em}
\noindent\fbox{%
    \parbox{0.9\linewidth}{%
    {\fontfamily{cmtt}\selectfont
    \small
        [CLS] \#\#\#  of Minnesota [SEP] Minneapolis [SEP] St. \#\#\# Park [SEP] \#\#\# Island Park \#\#\# Bell Museum [SEP] }
    }%
}
\vspace{0.2em}
\end{center}

In addition, we hypothesize that given common spatial co-occurrence patterns in the real world, one can use neighboring geo-entities to predict the name of a geo-entity. Therefore, we propose and incorporate a masked entity prediction (MEP) task, which randomly masks all subtokens of an entity name in a pseudo sentence. For MEP, \modelname relies on the masked entity's spatial relation to neighboring entities to recover the masked name. The block below shows an example of the masked input for MEP. 

\begin{center}
\vspace{0.1em}
\noindent\fbox{%
    \parbox{0.9\linewidth}{%
     {\fontfamily{cmtt}\selectfont
     \small
        [CLS] University of Minnesota [SEP] Minneapolis [SEP] \#\#\#  \#\#\#  \#\#\# [SEP] Bloom Island Park [SEP] Bell Museum [SEP] }
    }%
}
\vspace{0.2em}
\end{center}

Both MLM and MEP have a masking rate of 15\% (without masking \texttt{[CLS]} and \texttt{[SEP]}). The sequence position and spatial coordinates are not masked. 

\begin{figure}[h]
  \begin{center}
	\includegraphics[width=.85\linewidth]{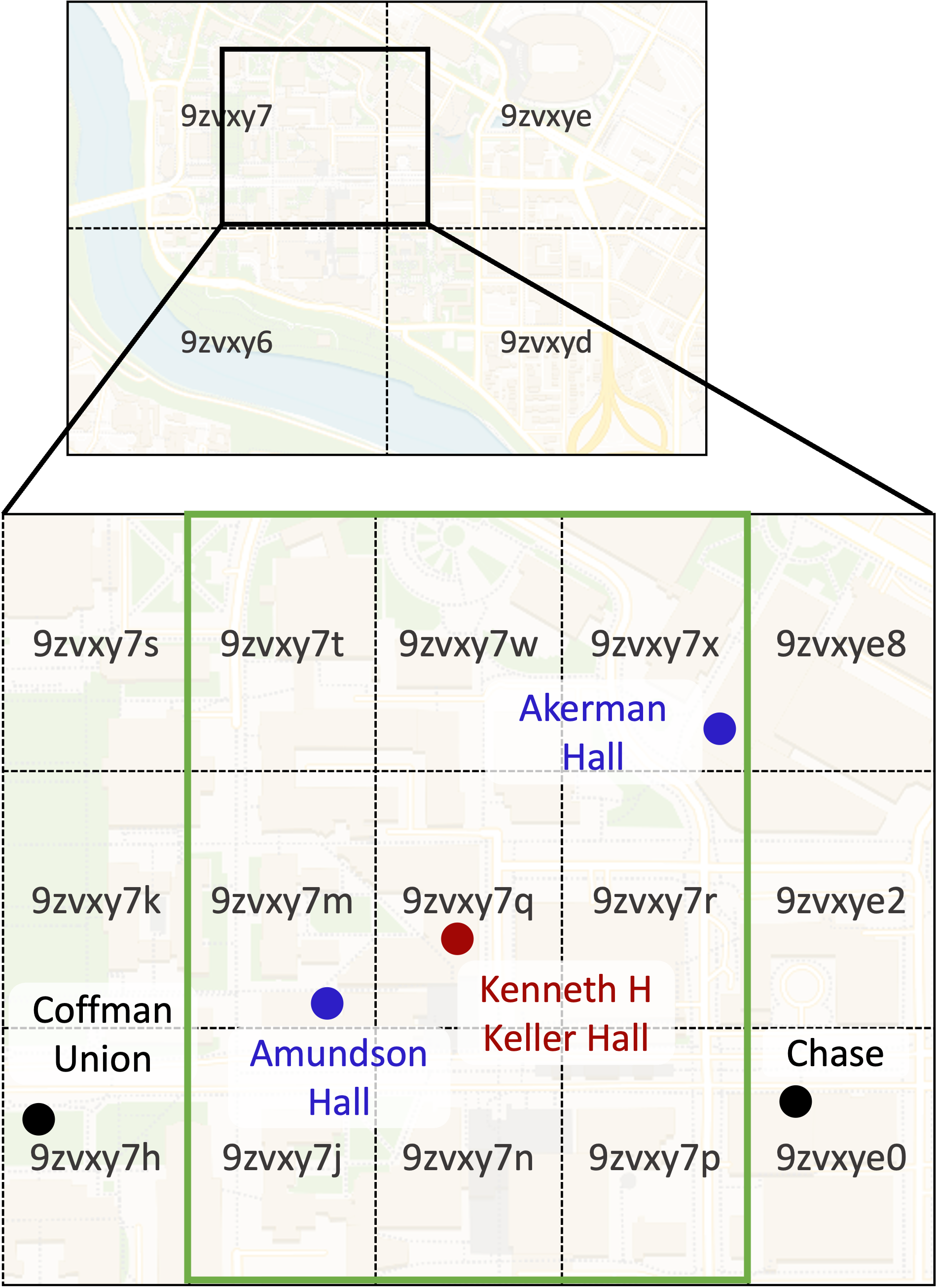}
  \end{center}
  \caption{Example of the Geohash representations with a pivot entity named `Kenneth H Keller Hall' and its neighboring entities from OpenStreetMap. (\tikz\draw[myred,fill=myred] (0,0) circle (.5ex);: pivot entity; \tikz\draw[myblue,fill=myblue] (0,0) circle (.5ex);: neighboring entities; \tikz\draw[black,fill=black] (0,0) circle (.5ex);: entities not within the nearest nine grids). The upper hash grids have a string length of six, and the lower ones have a length of seven.  }
  \label{fig:geohash}
  \vspace{-1em}
\end{figure}

\stitle{Pretraining Data}
We construct a large training set from OSM covering the City of London and the State of California. Since OSM is crowd-sourced, we clean the raw data by removing non-alpha-numeric place names and geo-entities that do not have a place name or geocoordinates. We randomly select geo-entities as pivots and construct pseudo sentences as described in \Cref{sec:overall}. 

To efficiently find the neighboring entities from a pivot sorted by distances, we leverage the Geohash algorithm introduced by Gustavo Niemeyer in 2008. We first generate the Geohash string for geo-entities on OpenStreetMap, which encodes a location into a string of a maximum length of 20 with base 32. For example, a city `\textit{Madelia}' with the geolocation (44.0508° N, 94.4183° W) has a Geohash string of `9zufe7nwjefpjksty0m8'. The length of the hash string is associated with the size of the geographic area that the string represents. The first character divides the entire Earth's surface into 32 large grids, and the second character further divides one large grid into 32 grids. Thus, by comparing the leading characters of two hash strings, we can estimate the spatial proximity of two locations without calculating the exact distance between them. We use the hash strings to first filter out the geo-entities guaranteed to be outside the spatial context radius and keep the neighbors within the nearest nine hash grids for distance calculation. The use of Geohash reduces the computation time when constructing the pseudo sentences. \Cref{fig:geohash} shows an example of Geohash representations. Given the pivot `Kenneth H Keller Hall', we can calculate its hash string of different lengths to quickly retrieve the neighboring entities within a desired range (i.e., the green box).


This way, we generate 69,168 and 148,811 pseudo sentences in London and California, respectively,
leading to a pretraining corpus of around 150M words.
We use all of them for pretraining \modelname with MLM and MEP. 

\subsection{Inference}\label{sec:inference}
The pretrained \modelname can already generate spatially varying contextualized representations for geo-entities. The representations support various downstream applications, including contextualized geo-entity classification and similarity-based inference tasks, such as geo-entity typing and linking. 

\stitle{Contextualized Geo-entity Classification} Classification of a geo-entity is crucial for recognizing and resolving geospatial concepts and the grounding of geo-entities in various data sources. Here, we use geo-entity typing as an example task to demonstrate the effectiveness of contextualized geo-entity representations. In this task, we aim to predict the geo-entity's semantic type (e.g., transportation and healthcare). We stack a softmax prediction head on top of the final-layer hidden states (i.e., geo-entity embeddings) to perform the classification.

\stitle{Similarity-based Inference} Integrating multi-source geographic datasets often involves finding the top $K$ nearest geo-entities in some representation space. One example task, which we refer to as geo-entity linking, is to link geo-entities from a 
geographic information system (GIS) oriented dataset to graph-based knowledge bases (KBs). In such a task, we can directly use the contextualized geo-entity embeddings from \modelname and calculate the embedding similarity based on some metrics, such as the cosine distance, to match the corresponding geo-entities from separate data sources.
\section{Experiments}
We evaluate \modelname on supervised geo-entity typing and unsupervised geo-entity linking tasks (\Cref{sec:setup}-\Cref{sec:linking_results}). We also investigate how \modelname performs under various common characteristics of geographic data sources (e.g., entity omission from cartographic generalization) and how critical technical components and their parameters affect the performance (\Cref{sec:analysis}). 
\subsection{Experimental Setup}\label{sec:setup}

\stitle{Datasets} \label{sec:dataset}
Open Street Map (OSM)\textsuperscript{\ref{ft:osm}} is a large-scale geographic database, and it is widely used in many popular map-related services. For \emph{supervised geo-entity typing}, we randomly select 25,872 and 7,726 pivot geo-entities together with their neighboring entities from point features of nine semantic types on OSM in the State of California and the City of London, respectively (\Cref{tab:typing_data_support}). Each geo-entity is associated with a name and its geocoordinates. We use 80\% of the data in both regions for training and 20\% for testing. The task is to predict the OSM semantic type for each geo-entity. (\Cref{sec:typing_results}).

For \emph{unsupervised geo-entity linking}, we randomly select 14 scanned historical maps in California from the United States Geological Survey (USGS) Historical Topographic Maps Collection. We manually transcribe text labels in these maps to generate 154 geo-entities, each containing a name (from the text label) and the image coordinates (text label's pixel location in the scanned map) The task is to find the corresponding Wikidata geo-entity\footnote{Entities with the coordinate location attribute (\url{https://www.wikidata.org/wiki/Property:P625})} for each USGS geo-entity with only their spatial relations from pixel locations. Note that the geocoordinates of geo-entities in the scanned maps are unknown. We select 15K Wikidata geo-entities having an identical name to one of the USGS geo-entities\footnote{Two geo-entities can have the same name (e.g., Los Angeles, CA vs. Los Angeles, TX)} and then randomly add another 30K Wikidata geo-entities to construct the final Wikidata dataset of 45K geo-entities. For the ground truth, we manually identify the matched Wikidata geo-entity for each USGS geo-entity.

For geo-entity linking task, the annotation details are described below. We hire three undergraduate students (not the co-authors) to transcribe text labels on the USGS maps. They draw bounding boxes/polygons around text labels and transcribe the text string to indicate geo-entity names and their locations. The results are verified and corrected by another undergraduate student and one Ph.D. student. Then, we run an automatic script to find the potential corresponding geo-entity in Wikidata by matching their names. The collected Wikidata URIs are noisy, and we have one Ph.D. student verify the URIs by marking them as True or False linking. The verification results are randomly further reviewed by a senior digital humanities scholar specialized in the representation and reception of historical places.

\tablesupport

\tabletypingresults

\stitle{Model Configuration}
We create two variants of \modelname, namely \modelnamens\modelsub{base} and \modelnamens\modelsub{large} with weights and tokenizers initialized from the \texttt{uncased} versions of the BERT\modelsub{base} and BERT\modelsub{large}, respectively. During pretraining, we mix the masked instances for MLM and MEP tasks in the same ratio. We primarily use the same optimization parameters as in \cite{devlin-etal-2019-bert} and train the model with the AdamW optimizer. We save the model checkpoints every 2K iterations. The learning rate is $5 \times 10^{-5}$ for \modelnamens\modelsub{Base} and $1\times 10^{-6}$ for \modelnamens\modelsub{Large} for stability. Distance normalization factor $Z$ is 0.0001, and distance separator \texttt{DSEP} is 20. The batch size is 12 to fit in one NVIDIA RTX A5000 GPU with 24GB memory.

\subsection{Supervised Geo-Entity Typing}\label{sec:typing_results}

\begin{figure}[t]
  \begin{center}
	\includegraphics[width=\linewidth]{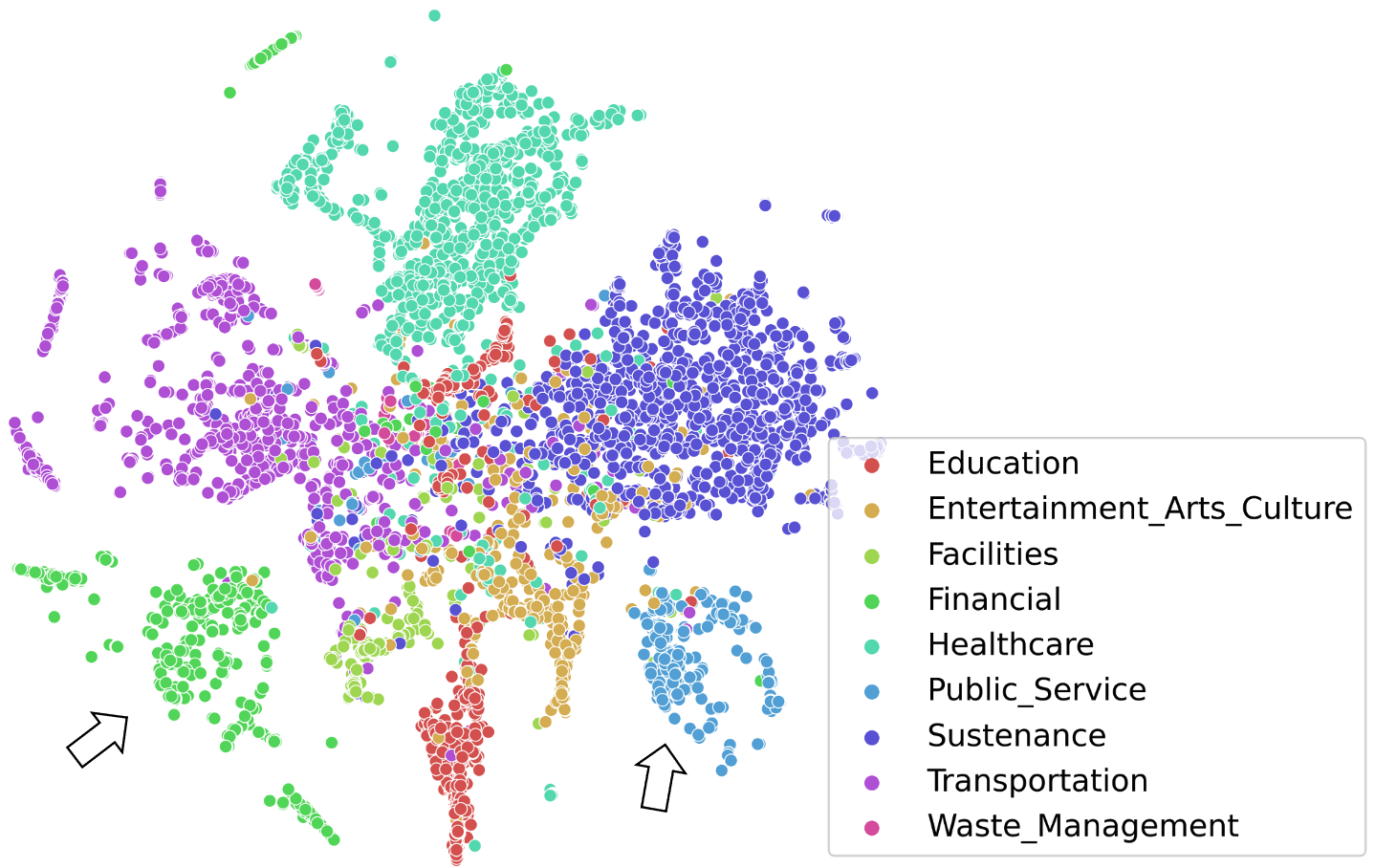}
  \end{center}
  \vspace{-0.5em}
  \caption{TSNE visualization for entity features produced by \modelnamens\modelsub{Base}. The color indicates the ground-truth labels. \textit{Financial} and \textit{Public\_Service} (pointed by the arrows) are well separated from other classes which conforms to the high F1 scores in \Cref{tab:typing_result} (best viewed in color).}
  \label{fig:tsne}
  \vspace{-0.5em}
\end{figure}


\stitle{Task Description} We tackle geo-entity typing as a supervised classification task using the modified \modelname architecture in \Cref{sec:inference} and finetune the model end-to-end by optimizing the cross-entropy loss. Specifically, the training set of this task contains $\{(g_i, SC(g_i), y_i)\}_{i=1}^N$, where $g_i$ is the pivot entity; $y_i$ is its OSM semantic type label; $SC(g_i)$ contains its neighboring entities in the nearest nine hash grids. Note that during training, the model does not use the neighboring geo-entities' OSM semantic types. We report the F1 score for each class (i.e., an OSM semantic type) and the micro-F1 for all samples in the test set.


\stitle{Model Comparison} We compare \modelname with several strong PLMs, including BERT \cite{devlin-etal-2019-bert}, RoBERTa~\cite{liu2019roberta}, SpanBERT~\cite{joshi-etal-2020-spanbert}, LUKE \cite{yamada-etal-2020-luke}, and SimCSE ~\cite{gao-etal-2021-simcse}. For all these models, we use the pretrained weights with the \texttt{uncased} version. For SimCSE, we use both the BERT and RoBERTa versions with the weights obtained from unsupervised training. We further finetune these baseline LMs end-to-end on the same pseudo-sentence dataset as \modelname (\Cref{tab:typing_data_support}) with a softmax prediction head appended after the baseline model. Note that these baseline models only use the token and position embeddings but not the spatial coordinate embeddings.

\stitle{Result Analysis} \Cref{tab:typing_result} summarizes the geo-entity typing results on the OSM test set with finetuned baselines and \modelname. The last column shows the micro-F1 scores (per-instance average). The remaining columns are the F1 scores for individual classes. For all baseline models, BERT\modelsub{Base} and SimCSE\modelsub{BERT-Large} (BERT\modelsub{Large} as a close second) have the highest F1 in the base and large groups, respectively. In addition, \modelname shows the best performance over the existing context-aware LMs for both base and large groups. We hypothesize that  \modelname's advantage comes from 1) the pretraining tasks, especially MEP, and 2) the spatial coordinate embeddings. Additionally, compared to its backbone, BERT, \modelname shows performance improvement in almost all classes for both base and large models. The results demonstrate that \modelname can effectively contextualize geo-entities with their spatial context.

We can also observe that the \textit{Financial} class has high F1 scores for all models. The reason is that when entity names are strongly associated with the semantic type (e.g., U.S. \underline{Bank}), pretrained LMs could produce meaningful entity representation even without the spatial context. However, the typing performance can still be further improved after encoding the spatial context with \modelname. \Cref{fig:tsne} visualizes the geo-entity features from \modelnamens\modelsub{Base}. It shows that \textit{Financial} and \textit{Public\_Service} have the most separable features from other classes. The non-separable region in the middle is mainly due to the similar spatial context and semantics of geo-entities (e.g. ``nursing home" can be close to both \textit{Facilities} and \textit{Healthcare}).




\subsection{Unsupervised Geo-Entity Linking} \label{sec:linking_results}
\stitle{Task Description} 
We define the problem as follows. For a query set $Q =  \{(q_i, SC(q_i))\}_{i=1}^{|Q|}$, the goal is to find the corresponding geo-entity for each $q_i$ in the candidate set $C =  \{(c_i, SC(c_i))\}_{i=1}^{|C|}$ (typically $|C| \gg |Q|$). Here, $Q$ and $C$ are the USGS and Wikidata geo-entities (\Cref{sec:dataset}). $SC(q_i)$ contains $q_i$'s nearest 20 geo-entities, and $SC(c_i)$ includes $c_i$'s neighboring geo-entities within a 5km radius.\footnote{Because $Q$ is not georeferenced, we use the top K nearest neighbors instead of pixel distances.} For this task, we use the evaluation metrics of Recall@K and Mean-Reciprocal-Rank (MRR).


\stitle{Model Comparison} We use the same baseline models as in the typing task without finetuning. 

\tablelinkingresults

\stitle{Result Analysis} \Cref{tab:linking_result} shows the geo-entity linking results. Large models perform better than their base counterparts for both baseline models and \modelname. In particular, among the baseline models, SimCSE\modelsub{BERT} has the highest score for MRR and R@1. \modelname has significantly higher R@5 and R@10 than SimCSE. Also, \modelname outperforms all baselines on MRR, R@5, and R@10. For both base and large versions, \modelname outperforms its backbone model BERT. The difference between BERT and \modelname shows the effectiveness of the spatial coordinate embeddings and domain adaptation using MLM and the proposed MEP. Also, SpanBERT shows the worst performance among all models. This could be that SpanBERT utilizes masked random spans in pretraining, which does not work well for pseudo sentences of geo-entity names. In contrast, \modelname's MEP specifically masks tokens from the same geo-entity and could effectively contextualize geo-entity names distributed in the 2D space with the spatial coordinate embeddings.



%

\subsection{Spatial Context and Ablation Study}\label{sec:analysis}

\stitle{Impact of Entity Omission for Linking} We analyze how \modelname handles different USGS map scales for geo-entity linking with entity omission from cartographic generalization (e.g., some entities only exist in certain scales). Here \modelname performs the best for the 1:125K maps (30-CA) (\Cref{tab:linking_result_scale}). Our hypothesis is that the 1:125K maps contain denser geo-entities than the test maps at other scales. When selecting the top K nearest neighbors for a pivot in other map scales, some neighbors could be far away and would not provide relevant context.

\tablelinkingscaleresults

\begin{figure}[t]
  \begin{center}
	\includegraphics[width=0.96\linewidth]{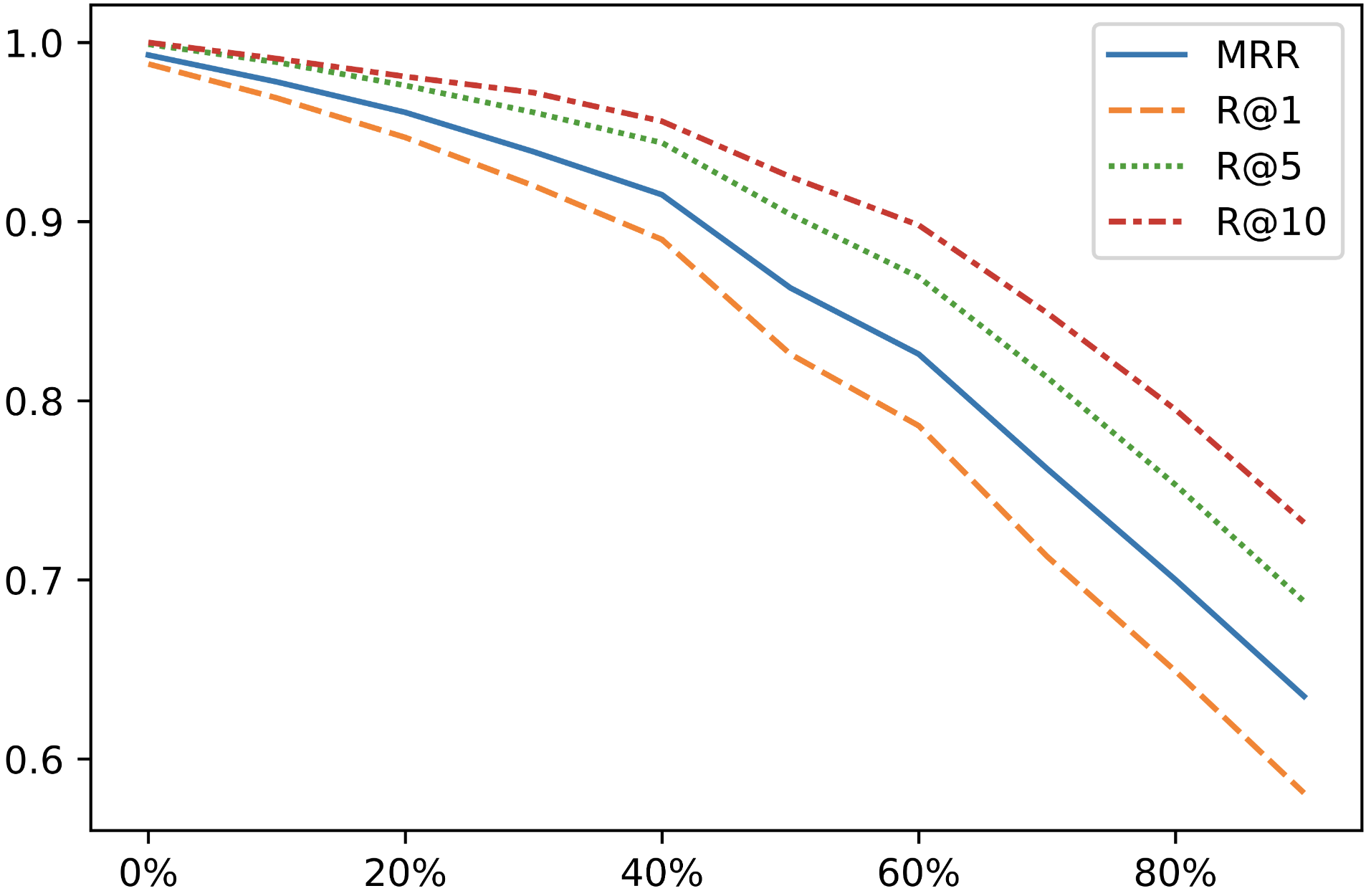}
  \end{center}
  \vspace{-0.5em}
  \caption{Geo-entity feature similarity with a decreasing neighbor density. X-axis indicates the percentage of the neighbors removed for the pivot entities. }
  \label{fig:line_plot}
  \vspace{-0.5em}
\end{figure}

Since the entity omission criteria of the USGS maps are unknown and Wikidata mostly contain important landmark geo-entities, we also simulate omission using the OSM dataset by gradually removing random neighbors of a pivot entity within a fixed neighborhood. We test this scenario for geo-entity linking to see if the same pivot entity would have similar representations from decreasing neighbor densities (i.e., the original set of neighbors vs. after random omission at varying rates). We use the same evaluation metric as in \cref{sec:linking_results}, as they reflect the similarity of the learned geo-entity representation (\Cref{fig:line_plot}). The linking scores decrease with additional entities removed, indicating the similarity between the same pivot entity's representations learned from the original data and the omitted data declines when omission percentage becomes larger. The elbow point is at about 30\%, showing that \modelname is reasonably robust if the percentage of the removed neighbors is within 30\%.

\stitle{Impact of Pseudo Sentence Length}
We analyze the impact of pseudo sentence length on geo-entity typing by varying the number of neighboring entities. \Cref{tab:typing_ablation} shows the results of \modelnamens\modelsub{Base} and \modelnamens\modelsub{Large} with an increasing number of neighbors. We observe that as \#Neighbor increases, the micro-F1 score increases accordingly, as expected. The increment is more evident at the beginning, illustrating that the spatial context helps contextualize the pivot entity, especially in a local neighborhood. The benefit of adding additional neighboring entities decreases as distance increases beyond a certain point. This conforms to the well-studied spatial autocorrelation. 

\tabletypingablation

\stitle{Effect of Spatial Coordinate Embedding}
We train two additional \modelname variants that do not include the spatial coordinate embedding during MLM and MEP pretraining and use them for geo-entity linking.  
For MRR, \modelnamens\modelsub{Base} and \modelnamens\modelsub{Large} drop from 0.515 to 0.458 and from 0.537 to 0.478 compared to their original \modelname version. Also, the variants have lower recall scores compared to their original \modelname version. The most significant drop in the recall is on R@1 from 0.383 to 0.283 for \modelnamens\modelsub{Large}. The results demonstrate the effectiveness of the spatial coordinate embedding.



\section{Related Work}



\stitle{Pretrained Language Models}
PLMs have been the dominant paradigm for language representation.
Following the success of MLMs \cite{devlin-etal-2019-bert,liu2019roberta} and autoregressive PLMs \cite{peters-etal-2018-deep,radford2019language},
more recent work has extended the pretraining process with more tasks or learning objectives for span prediction \cite{joshi-etal-2020-spanbert}, cross-encoders \cite{reimers-gurevych-2019-sentence}, contrastive learning \cite{gao-etal-2021-simcse}, and massively multi-task learning \cite{raffel2020exploring}.
To support entity representation, several approaches propose to perform mention detection \cite{yamada-etal-2020-luke}, incorporate mention memory cells \cite{de2021mention}, or injecting structural knowledge representations \cite{wang-etal-2021-k,zhang-etal-2019-ernie,peters-etal-2019-knowledge,zhou-etal-2022-prix}.
Due to the large body of work, we refer readers to recent surveys summarizing this line of work \cite{qiu2020pre,wei2021knowledge}.

To extend the use of PLMs beyond language, much exploration has also been conducted to represent other modalities.
For example, a significant amount of vision-language models \cite{kim2021vilt,zhai2022lit,li-etal-2020-bert-vision,lu2019vilbert} have been developed by jointly pretraining on co-occurring vision and language corpora,
and are in the support of grounding \cite{zhang-etal-2020-learning-represent,chen2020uniter}, generation \cite{pmlr-v139-cho21a,yu2022scaling} and retrieval tasks \cite{zhang-etal-2021-visually-grounded,li2020oscar} on the vision modality.
Other PLMs capture semi-structure tabular data by linearizing table cells \cite{herzig-etal-2020-tapas,yin-etal-2020-tabert,iida-etal-2021-tabbie} or incorporating structural prior in the attention mechanism \cite{wang-etal-2022-robust,trabelsi2022strubert,eisenschlos-etal-2021-mate}. To the best of our knowledge, none of the prior studies have extended pretrained LMs to geo-entity representation, nor do they support a suitable mechanism to capture the geo-entities' spatial relations, which do not have structural prior. This is exactly the focus of our work.


\stitle{Domain-specific Language Modeling} 
Another line of studies has been conducted to adapt PLMs to specific domains typically by pretraining on domain-specific corpora.
For example, in the biomedicine domain, a series of models \cite{lee2020biobert,peng-etal-2019-transfer,alsentzer-etal-2019-publicly,phan2021scifive} have been developed by training PLMs on corpora derived from PubMed. 
Similarly, PLMs have been trained on corpora specific to software engineering \cite{tabassum-etal-2020-code}, finance \cite{liu2021finbert}, and proteomics \cite{zhou2020mutation} domains. In this context, \modelname represents a pilot study in the geographical domain by allowing the PLM to learn from spatially distributed text.

\section{Conclusion}
This paper presented \modelname (\logo), a language model trained on geographic datasets for contextualizing geo-entities. \modelname utilizes a novel spatial coordinate embedding mechanism to capture spatial relations between 2D geo-entities and linearizes the geo-entities into 1D sequences, compatible with the BERT-family structures. The experiments show that the general-purpose representations learned from \modelname achieve better or competitive results on the geo-entity typing and geo-entity linking tasks compared to SOTA pretrained LMs. We plan to evaluate \modelname on additional related tasks, such as geo-entity to natural language grounding. Also, we plan to extend \modelname to support other geo-entity geometry types, including lines and polygons. 

\section*{Limitations}
The current model design only considers points but not polygon and line geometries, which could also help provide meaningful spatial relations for contextualizing a geo-entity. Training of \modelname also requires considerable GPU resources which might produce environmental impacts.

\section*{Ethical Consideration}
\modelname was evaluated on the English-spoken regions, so the model and results could have a bias towards these regions and their commonly used languages. Replacing the backbone of \modelname with a multi-lingual model and training \modelname with diverse regions could mitigate the bias.

\section*{Acknowledgement}
We thank the reviewers for their insightful comments and suggestions. We thank Dr. Valeria Vitale for her valuable input. This material is based upon work supported in part by NVIDIA Corporation, the National Endowment for the Humanities under Award No. HC-278125-21 and Council Reference AH/V009400/1, the National Science Foundation of United States Grant IIS 2105329, a Cisco Faculty Research Award (72953213), and the University of Minnesota, Computer Science \& Engineering Faculty startup funds.

\newpage

\bibliography{anthology,custom}
\bibliographystyle{acl_natbib}


\end{document}